\documentclass[11pt]{article}

\usepackage[utf8]{inputenc} 
\usepackage{hyperref}       
\usepackage{url}            
\usepackage{booktabs}       
\usepackage{amsfonts}       
\usepackage{nicefrac}       
\usepackage{microtype}      
\usepackage{lipsum}

\usepackage{enumerate}
\usepackage[english]{babel}
\usepackage{multicol}
\usepackage[dvipsnames]{xcolor}
\usepackage{color}
\usepackage{subcaption}

\usepackage{authblk}

\usepackage{amssymb}
\usepackage{graphicx}
\usepackage{float}

\newcommand{\argmin}{\operatornamewithlimits{argmin}}

\usepackage{glossaries}
\usepackage[ruled,vlined]{algorithm2e}

\title{Learning chordal extensions\thanks{Submitted to Journal of Global Optimization.}}

\author{Defeng Liu, Andrea Lodi, Mathieu Tanneau}

\affil{
    \texttt{\{defeng.liu, andrea.lodi, mathieu.tanneau\}@polymtl.ca}
}
\affil{Canada Excellence Research Chair\\ Polytechnique Montr\'eal }

\date{\vspace{-5ex}}

\begin{document}
\maketitle

\begin{abstract}
    A highly influential ingredient of many techniques designed to exploit sparsity in numerical optimization is the so-called chordal extension of a graph representation of the optimization problem.
    The definitive relation between chordal extension and the performance of the optimization algorithm that uses the extension is not a mathematically understood task.
    
    For this reason, we follow the current research trend of looking at Combinatorial Optimization tasks by using a Machine Learning lens, and we devise a framework for learning elimination rules yielding high-quality chordal extensions. As a first building block of the learning framework, we propose an on-policy imitation learning scheme that mimics the elimination ordering provided by the (classical) minimum degree rule. 

    The results show that our on-policy imitation learning approach is effective in learning the minimum degree policy and, consequently, produces graphs with desirable fill-in characteristics.
\end{abstract}

\section{Introduction}
\label{sec:intro}
A simple undirected graph $G = (V, E)$ is \emph{chordal} if, for every cycle $c$ of length at least four, there exists an edge $e \in E$ that connects two non-consecutive vertices of $c$.
A \emph{chordal extension} of a graph $G$ is a chordal graph $H$ such that $G$ is a sub-graph of $H$, i.e., one can obtain $H$ by adding edges to $G$.
A practical way of constructing chordal extensions is via \emph{graph elimination} \cite{Vandenberghe2015}, which consists in sequentially eliminating the nodes of the graph.
At each step, a node $v$ is selected, new edges are inserted so as to make the neighbors of $v$ into a clique, then $v$ is removed (i.e., eliminated).
This process is repeated until all nodes have been eliminated, and one obtains a chordal extension by adding to the original graph all edges that were inserted in the process.
The order in which nodes were eliminated is thereby called an \emph{elimination ordering}.

This work focuses on the role of chordal extensions and graph elimination within optimization frameworks.
Indeed, there is a direct connection between chordal extensions, which are typically computed via graph elimination, and a number of classical sparsity-exploiting techniques \cite{Wright1997,Vandenberghe2015}.
In particular, we seek to devise a framework for \emph{learning} elimination rules that yield high-quality chordal extensions, as we illustrate below.

Our first motivating example is the computation of a fill-reducing ordering for sparse Cholesky factorization, a process that reduces to computing an elimination ordering \cite{Fomin2015,Vandenberghe2015}.
Crucially, Cholesky factorization underlies most implementations of interior-point algorithms for linear programming \cite{Wright1997,Bixby2002}, and the choice of ordering can have a major impact on the method's performance \cite{Rothberg1998}.
Similarly, chordal graphs form the basis of \emph{chordal decomposition} techniques to exploit sparsity in semi-definite programming (SDP) problems, see, e.g., \cite{Agler1988,Majumdar2019}.
Specifically, a single, dense, semi-definite constraint, can be decomposed into several smaller, yet coupled, semi-definite constraints.
This reformulation also reduces to computing a chordal extension, and can dramatically improve the performance of both interior-point and first-order methods on large problems \cite{Zheng2019,Majumdar2019}.
More generally, a similar approach can be leveraged in linear conic optimization and convex optimization, see, e.g., \cite{Vandenberghe2015}.

Historically, efforts have focused on computing minimum chordal extensions, i.e., chordal extensions with a minimum number of additional edges \cite{Rothberg1998,Bergman2019}, which has been proven to be NP-complete \cite{Yannakakis1981}.
This fostered the development of fast and efficient heuristics such as minimum degree \cite{George1989} and nested dissection \cite{NestedDissection} orderings.
State-of-the-art implementations of these methods are routinely used in most optimization and sparse linear algebra software, where they tackle problems with up to millions of variables.
Nevertheless, which chordal extension is computed can significantly impact the subsequent performance of the optimization algorithm.
Therefore, it is natural to seek a ``best" chordal extension, i.e., one that maximizes performance.

Recently, the use of \emph{Machine Learning} (ML) in \emph{Combinatorial Optimization} (CO) became a popular research area with quite a number of contributions investigating many angles of such a connection. On the one hand, some research has been devoted to solve CO problems by ML, i.e., to devise new heuristic algorithms that perform the end-to-end learning of the solution of a CO problem. On the other hand, ML has been used to tackle some tasks within CO algorithms and software for which modern statistical learning has chances to improve the current performances, either because the known way of performing those tasks is computationally heavy or because they are poorly understood from the mathematical standpoint. The interested reader is referred to \cite{bengio2018} for a methodological survey on this new research area.

Our work does not follow the first direction outlined in the previous paragraph. Indeed, the goal of this paper is \emph{not} to compete with existing state-of-the-art heuristics for graph elimination.
First, these heuristics leverage decades of development and clever engineering, and have been optimized for fast runtime and good solution quality. Second, and more importantly, the aim of our work is to eventually gain deeper understanding of the relation between the characteristics of a chordal extension and the behavior of optimization algorithms and software, a topic of interest in its own right and whose mathematical knowledge is currently insufficient.
In order to achieve this ultimate goal, our present contributions are 1) to propose a mathematical framework for the problem of learning elimination orderings, and 2) to provide methodological and practical insights on the learning process itself. As a byproduct, a better understanding of what ``good" chordal extensions look like (for a specific task) can lead to an easier and more direct customization of elimination orderings to sets of similar graphs.

The rest of the paper is organized as follows.
In Section \ref{sec:notation}, we introduce some relevant definitions and concepts.
In Section \ref{sec:learning}, we present our methodology for learning chordal extensions. 
In Section \ref{sec:experiments}, we report on numerical experiments. Section \ref{sec:further} gives further discussion and Section \ref{sec:conclusion} concludes the paper.

\section{Basic notations and concepts}
\label{sec:notation}
In this section, we review some notations and concepts used in the remainder of the paper. Section \ref{sec:graphs} introduces basic notations and definitions of graphs and Section \ref{sec:mdp} introduces a commonly used model for sequential decision problems. In Sections \ref{sec:statistical_learning} and \ref{sec:imitation}, we briefly go through some Machine Learning concepts, in order to help the reader be familiar with relevant ML methods and properly locate the methodology we propose in Section \ref{sec:learning}.

In all that follows, for an arbitrary set $S$, we use the notation $\mathfrak{P}(S)$ to denote the set of all probability distributions over $S$.

\subsection{Graph-theoretic notations}
\label{sec:graphs}

In this paper, all considered graphs are simple, undirected graphs.
A graph is denoted by $G = (V, E)$, where $V$ (resp. $E$) denotes the set of its nodes (resp. its edges).
For an edge $e = (v, w)$, we say that $e$ is \emph{incident} to $v$ and $w$, that $v, w$ are the \emph{extremities} of $e$, and that $v, w$ are \emph{adjacent}.
The neighborhood of $v$, denoted by $\mathcal{N}_{G}(v)$, is defined as the set of nodes that are adjacent to $v$.

The degree of node $v$ in $G$ is denoted by $\delta_{G}(v)$.
We say that node $v$ is of \emph{minimum degree} if $\delta_{G}(v) \leq \delta_{G}(w)$ holds for all nodes $w \in V$.
Note that several nodes of minimum degree may exist.

In what follows, we will drop the subscript $G$ whenever context is sufficiently clear, and write for example $\delta(v)$ rather than $\delta_{G}(v)$.

\subsection{Markov Decision Processes}
\label{sec:mdp}

\emph{Markov Decision Processes} (MDPs) \cite{howard1960dynamic} are wildly used to formulate sequential decision problems.
An MDP is characterized by a set of possible \emph{states} $\mathcal{S}$, a set of possible actions $\mathcal{A}$, the dynamics of the system, and a cost function.

Formally, an MDP is defined by a tuple $(\mathcal{S}, \mathcal{A}, P, c)$, where $P$ encodes the system's dynamics, and $c$ encodes the cost function.
For any state-action pair $(s, a) \in \mathcal{S} \times \mathcal{A}$, $P(s, a)$ is a probability distribution over the state space $\mathcal{S}$.
That is, if one takes action $a$ in state $s$, the next observed state $s'$ will be sampled from the distribution $P(s, a) \in \mathfrak{P}(\mathcal{S})$.
Finally, the cost of executing action $a$ in state $s$ is denoted by $c(s, a)$.

Given an MDP $(\mathcal{S}, \mathcal{A}, P, c)$, a \emph{policy} is a decision rule for selecting an action $a$, given a current state $s$.
Specifically, a policy $\pi$ is a function
\begin{align}
    \pi : \mathcal{S} & \longmapsto \mathfrak{P}(\mathcal{A})\\
         s &\longrightarrow \pi(s),
\end{align}
and the next action $a$ is sampled from $\pi(s) \in \mathfrak{P}(\mathcal{A})$.
If a policy $\pi$ maps each state to a \emph{single} action, i.e., if $\pi(s)$ is a degenerate distribution for every $s \in \mathcal{S}$, then $\pi$ is called a \emph{deterministic policy}; otherwise it is called a \emph{stochastic policy}.
For the remainder of this paper, we will only consider stochastic policies, and we define the expected \emph{immediate cost} for policy $\pi$ in state $s \in \mathcal{S}$ as
\begin{align}
    C_{\pi}(s) &= \mathbb{E}_{a \sim \pi(s)} \left[
        c(s, a)
    \right].
\end{align}

A \emph{trajectory} is a sequence of state-action pairs $\big( (s_{0}, a_{0}), (s_{1}, a_{1}), ...\big)$ where $s_{t+1}$ is sampled from $P(s_{t}, a_{t})$.
In this paper, all trajectories will always be finite, although they may be of arbitrary length.
We say that a trajectory  is sampled from a policy $\pi$ if each action $a_{t}$ is sampled from $\pi(s_{t})$.
The total cost along a trajectory is then given by
\begin{align}
    \sum_{t \geq 0} c(s_{t}, a_{t}),
\end{align}
which is always finite since we only consider finite trajectories.

Finally, for a distribution of initial states $\mathcal{D}_{0} \in \mathfrak{P}(\mathcal{S})$, we define the \emph{expected total cost} of a policy $\pi$ as
\begin{align}
    \label{eq:expected_cost}
    C^{tot}_{\pi} &= \mathbb{E}_{s_{0} \sim \mathcal{D}_{0}} \left(
        \mathbb{E}_{\pi} \left[
            \sum_{t \geq 0} c(s, a)
        \right]
    \right),
\end{align}
where $\mathbb{E}_{\pi}$ denotes that trajectories are sampled from $\pi$.
Furthermore, $\mathcal{D}_{0}$ and $\pi$ induce a stationary distribution over states, which we denote by $\mathcal{D}_{\pi}$.
Thus, we define the \emph{expected average cost} of policy $\pi$ as
\begin{align}
    C^{avg}_{\pi} &= \mathbb{E}_{s \sim \mathcal{D}_{\pi}} \left[
            \mathbb{E}_{a \sim \pi(s)} \left( c(s, a) \right)
        \right]\\
        &= \mathbb{E}_{s \sim \mathcal{D}_{\pi}} \left[ C_{\pi}(s) \right].
\end{align}
In this work, we assume that the above two expectations are always finite; we make this mild technical assumption to ensure that the learning problems defined in Section \ref{sec:imitation} are well-defined.
For ease of reading, we also drop the explicit dependency of $C^{tot}_{\pi}$ and $C^{avg}_{\pi}$ on $\mathcal{D}_{0}$, since the latter will always be evident from the context.

\subsection{Standard statistical learning}
\label{sec:statistical_learning}
In statistical learning, the goal is to detect patterns from a set of observed data and make predictions about future data. We can formalize the standard statistical learning problem as follows. Given a variable space $\mathcal{Z}$ and a set of examples $\mathcal{D_Z} = \{z_1,z_2, \dots, z_m\}$ from the unknown distribution $\mathcal{P(Z)}$,
the task is to find a function $f$ over a family of functions $\mathcal{F}$, such that $f$ ``performs well" on $\mathcal{P(Z)}$. It is assumed that all the observed examples are drawn independent and identically distributed (i.i.d.) from the same distribution $\mathcal{P(Z)}$.

If a \emph{loss} function $\mathcal{L}: \mathcal{F} \times \mathcal{Z} \mapsto \mathbb{R}$ is specified to measure the performance of $f$, then the goal can be described as finding $\hat{f} \in \mathcal{F} $ that minimizes the \emph{expected  loss} with respect to  $\mathcal{P(Z)}$, i.e.,
\begin{align}
    \hat{f} = \argmin_{f \in \mathcal{F}} \mathbb{E}_{ z \sim \mathcal{P(Z)}}{\left[ \mathcal{L}(f, z)\right] }.
\end{align}
The learned function $\hat{f}$ is then used to predict future data. 

However, the expected loss cannot be computed exactly due to the fact that $\mathcal{P(Z)}$ is unknown. In practice, if a subset of examples $\mathcal{D_Z}$ sampled from $\mathcal{P(Z)}$ is available, a number of learning methods turn to minimize the \emph{empirical loss} on $\mathcal{D_Z}$. Then, $\hat{f}$ is obtained by solving 
\begin{align}
    \hat{f} = \argmin_{f \in \mathcal{F}} \frac{1}{m} \sum_{i=1}^{m}{ \mathcal{L}(f, z_i) }.
\end{align}

The differences in forms and contents of $\mathcal{Z}$, $\mathcal{F}$, $\mathcal{L}$ result in different learning tasks. Here, we only introduce \emph{supervised learning} that is the relevant task for the current state of our work.

 \paragraph{Supervised learning.}
 In supervised learning, the variable space $\mathcal{Z}$ consists of $\mathcal{X} \times \mathcal{Y}$, where $\mathcal{X}$ is the space of input variables and $\mathcal{Y}$ is the space of output variables. The family of functions $\mathcal{F}$ is a set of mappings $f: \mathcal{X} \mapsto \mathcal{Y}$. For any sample $(x, y) \in \mathcal{X} \times \mathcal{Y}$ , the loss function $\mathcal{L}$ measures the discrepancy between $f(x)$ and $y$. Ideally, the output $\mathcal{Y}$ can be in any form or intent. However, most tasks assume that $\mathcal{Y}$ is categorical or nominal. The former characterizes the task as classification, whereas the latter induces regression. 
 
\subsection{Imitation learning for sequential decision problems}
\label{sec:imitation}

\emph{Imitation learning} (IL) \cite{bagnell2007boosting,ratliff2009learning,ross2011reduction,schaal1999imitation} is an extension of supervised learning from problems satisfying i.i.d. assumption to sequential decision problems, see, e.g., \cite{pan2018agile,ross2010efficient,silver2008high}. 

In IL, the target policy learns its decision rule from an expert policy. 
More precisely, given the class of candidate policies $\Pi$, we seek to find a target policy $\pi \in \Pi$ that matches the expert policy $\pi^*$. The target and expert policy are often referred as the \emph{learner} and the \emph{expert}. The cost is defined by a loss function $\mathcal{L}(\pi,\pi^*)$, a measure of discrepancy between $\pi$ and $\pi^*$.  If a \emph{behavior policy} $\pi'$ is set to generate trajectories of states, the goal is to find a policy $\hat{\pi}$ that minimizes the expected loss with respect to the distribution of states induced by $\pi'$, namely
\begin{align}
    \hat{\pi} = \argmin_{\pi \in \Pi} \mathbb{E}_{ s \sim \mathcal{D}_{\pi'}}{\left[ \mathcal{L}\left(\pi(s), \pi^*(s)\right)\right]}.
\end{align}

In the literature, depending on the choice of behavior policy, IL algorithms can be divided into two classes: \emph{off-policy} and \emph{on-policy} methods. The off-policy methods are the ones with behavior policy $\pi'$ independent of the learner $\pi$. For instance, the supervised approach for imitation learning falls into this class. On the other hand, if the behavior policy depends on the learner, the algorithm becomes an on-policy method.
Data Aggregation (DAGGER)\cite{ross2011reduction} and its variants are examples of on-policy methods.
    
\paragraph{Supervised imitation learning.} The supervised approach for IL fixes the behavior policy as the expert, i.e., $\pi'=\pi^*$. Then the learner is trained under the distribution induced by the expert, given by
\begin{align}
    \hat{\pi} =  \argmin_{\pi \in \Pi} \mathbb{E}_{ s \sim \mathcal{D}_{\pi^*}}{\left[ \mathcal{L}\left(\pi(s), \pi^*(s)\right)\right]}.
\end{align}

\paragraph{Data Aggregation.} DAGGER is an iterative algorithm that improves the learner by executing a mixed behavior policy combined with the learner and the expert. 
At each iteration, the collected data will be aggregated into an accumulated dataset and the learner will be trained by all the data collected from previous iterations.


\section{Methodology}
\label{sec:learning}

In this section, we present our methodology for learning chordal extensions. In this work, we focus on how to learn elimination rules for graph elimination via imitation learning. More precisely, we propose an on-policy imitation learning scheme that mimics the elimination ordering provided by the ordering heuristic we choose.

\subsection{MDP formulation}
\label{sec:learning:mdp_ge}
    We begin by formulating graph elimination as a Markov decision process.
    First, the state space $\mathcal{S}$ is the set of simple undirected  graphs.
    Then, for a given graph $G = (V, E)$, the corresponding set of possible actions is identified by the nodes of the graph.
    Transitions are deterministic: if an action $a=v$ is performed in state $G$, i.e., if node $v$ is eliminated from graph $G$, then the new state is uniquely defined as the graph obtained from the elimination of node $v$.
    Note that the number of nodes decreases by one at each step.
    Hence, even though the initial graph may be of arbitrary size, trajectories are always finite.
    
    Thus, a policy $\pi$ maps a graph to a probability distribution over its set of nodes $V$.
    Therefore, if $V = \{1, ..., n\}$, then $\pi(G)$ is a $n$-dimensional non-negative vector, whose $i$-th coordinate  denotes the probability that node $i$ be eliminated.
    
    Finally, one may select a cost function according to the problem at hand, for example, the number of additional edges, i.e., fill-in.
    In that case, finding a policy that minimizes the expected total cost reduces to finding a policy that yields minimum chordal extensions.
    Rather than trying to minimize fill-in, which is an NP-hard problem for which efficient heuristics already exists, we adopt a more generic imitation learning scheme as discussed below.

\subsection{On-policy imitation learning}
\label{sec:learning:onpolicyil}
Although the goal of imitation learning is to find a learner that best matches the expert, any parameterized stochastic policy will inevitably have chances to make occasional mistakes by choosing an action different from the expert. In the supervised imitation learning approach, where the learner is only trained under the distribution of states induced by the expert, the learner may not be able to correct its behavior from deviations induced by its bad choice of action. 

Moreover, any possible state is observable in our problem setting and the expert is always accessible for querying any state. Therefore, we choose on-policy approach in order to be more robust to alleviate the deviation problem. Specifically, we set the learner as the behavior policy, i.e., $\pi' = \pi$. As a result, the goal is to find a policy $\hat{\pi}$ such that
\begin{align}
    \hat{\pi} =  \argmin_{\pi \in \Pi} \mathbb{E}_{ s \sim \mathcal{D}_{\pi}}{\left[ \mathcal{L}\left(\pi(s), \pi^*(s)\right)\right]}.
\end{align}

Since, in general, the expected loss cannot be computed analytically, we estimate it by sampling trajectories of states from finite dataset using $\pi$. Given a dataset $\mathcal{G}$ of $M$ graphs, we train the learner with our practical algorithm, namely \emph{one-step on-policy imitation learning}, as described in Algorithm \ref{alg:the_alg}.



\begin{algorithm}[H]
\SetAlgoLined
\label{alg:the_alg}
\KwIn{Instance Dataset $\mathcal{G} = \{ \textbf{g}_i \}_{i=1}^M $}
 Initialize $\pi_{\theta}$ to any policy in $\Pi$\;

 \For{$i = 1$ \textnormal{\textbf{to}} $N$} {
    \For {\textnormal{\textbf{each}} $g \in \mathcal{G} $} {
        Initialize $s$\;
        \For{$j = 1$ \textnormal{\textbf{to}} $T$} {
            $\theta \gets \theta - \alpha \nabla_{\theta} \mathcal{L}\left(\pi_{\theta}(s), \pi^*(s)\right)$\;
            $a \gets \pi_{\theta}(s)$\;
            Take action $a$, observe next state $s$\;
            }
        }
    }
 \Return $\pi_{\theta}$
 \caption{One-step on-policy imitation learning}
\end{algorithm}

The training proceeds as follows.
At every epoch $i$ (from $1$ to $N$), each instance $G=(V, E) \in \mathcal{G}$ is used to generate one complete trajectory (with length $T = |V|$), and each state (i.e., graph) in this trajectory represents a training data point.
The one-step updating is as follows.
At every transition, i.e., at every elimination step, the leaner $\pi_{\theta}$ is updated by taking a gradient step with respect to the loss $\mathcal{L}\left(\pi_{\theta}(s), \pi^*(s)\right)$, where $s$ is the current state.
An action $a$ is then sampled from the updated learner $\pi_{\theta}$, which yields the next state $s'$.
Note that, in this setting, the loss is computed by comparing the two distributions $\pi_{\theta}(s)$ and $\pi^{*}(s)$ directly.
We do so because 1) we know analytically both the learner and the expert, and 2) it allows to exploit information from the entire distributions rather than sampling and comparing individual actions.

    \paragraph{Expert.}
   We consider the minimum degree heuristic \cite{Markowitz1957,George1989} as the expert. At each step, the minimum degree selects a node of minimum degree to be eliminated.
    Ties are broken arbitrarily, i.e., if several nodes have minimum degree, then one is selected uniformly at random among them.
    Let us note that today's implementations include several additional features, such as smarter tie breaking or the simultaneous elimination of multiple nodes.

    In all that follows, we denote $\pi_{MD}$ the minimum-degree policy, i.e., for a given graph $G = (V, E)$, we have
    \begin{align*}
        \pi_{MD}(G) [v]= 
        \left\{
        \begin{array}{rl}
            \frac{1}{k} & \ \ \ \text{if $v$ is of minimum degree} \\
            0 & \ \ \ \text{otherwise }
        \end{array}
        \right.
        ,
    \end{align*}
    where $k$ is the number of nodes that have minimum degree.
	
    \paragraph{Learner parameterization.}
    Given that states are represented as graphs, with arbitrary size and topology, we propose to use \emph{graph neural networks} (GNNs) \cite{gori2005new,hamilton2017representation} to parameterize the learner.
    Indeed, GNNs is an expressive type of model to process graph-structured data and have been applied to a variety of representation learning tasks on graphs \cite{duvenaud2015convolutional,kipf2016semi,hamilton2017inductive,li2018combinatorial,gasse2019}.
    In GNN models, a graph is embedded according to the features of vertices and its topological structure.
    The encoding of a node is generated by propagating information from its neighborhood.
    One of the most appealing properties of GNNs is that it is size-and-order invariant to input data, i.e., it can process graphs of arbitrary size, and the ordering of the input elements is irrelevant.
    
    In our GNN architecture, the embedding of the graph in the $(l+1)$-th layer is computed by aggregating, for each node, the features of its neighborhood from the $l$-th layer, i.e.,
    \begin{align}
        \textbf{H}^{l+1} =f \left(\textbf{A} \textbf{H}^{l} \textbf{W}^l  + \textbf{I}_{n\times 1}\textbf{B}^l \right),
    \end{align}
   where $\textbf{A} {\in} \mathbb{R}^{n \times n}$ is the adjacency matrix of the graph, $\textbf{I}_{n\times 1}$ is the matrix of ones with size of $n \times 1$, $d^l$ and $d^{l+1}$ are the dimension of the features in layer $l$ and $(l+1)$, $\textbf{H}^{l+1} {\in} \mathbb{R}^{n \times d^{l+1}}$ and $\textbf{H}^l {\in} \mathbb{R}^{n \times d^{l}}$ are the embeddings of layer $(l+1)$ and $l$, $\textbf{W}^l {\in} \mathbb{R}^{d^{l} \times d^{l+1}}$ and $\textbf{B}^l {\in}  \mathbb{R}^{1 \times d^{l+1}}$ are the parameters in layer $l$, and $f(\cdot)$ specifies the activation function. For  $f(\cdot)$, we apply $Softmax$ in the output layer and $Relu$ in the rest.
    
    
    \paragraph{Loss function.} 
    Since the expert is the minimum degree heuristic, we can compute the exact $\pi_{MD}(s)$ given a state $s$ of a graph as shown before. To measure the distance between two distributions given by the learner and the expert respectively, we compute the Kullback-Liebler (KL) divergence \cite{kullback1997information} between $\pi_{MD}(s)$ and $\pi(s)$ as the loss.

\section{Numerical experiments}
\label{sec:experiments}

In this section, we report the details of our computational investigation. More precisely, Section \ref{sec:experiments:datasets} specifies the data generation and collection. In Section \ref{sec:experiments:settings}, we discuss the experimental setting and, finally, Section \ref{sec:experiments:results} reports the computational results.

\subsection{Data collection}
\label{sec:experiments:datasets}

We evaluate our approach on four different datasets, which comprise graphs that vary in size and structural characteristics.

\subsubsection{Erdos-Renyi graphs}

We first build two datasets of Erdos-Renyi graphs, a simple and well-known class of random graphs.
We use the notation $G(n, p)$ to denote a (random) Erdos-Renyi graph with $n$ nodes, and such that edges are selected with probability $p \in [0,1]$ independently of each other.
Note that, for given $n$ and $p$, $G(n, p)$ is a random variable whose realizations are graphs of size $n$.
While $n$ controls the size of the graph, $p$ controls its sparsity.

We form two datasets of Erdos-Renyi graphs: one of smaller graphs, denoted $ER_{S}$, and the other one of larger graphs, denoted $ER_L$.

Each graph in $ER_{S}$, is sampled from $G(n, p)$, where $n$ is drawn uniformly between  $100$ and $300$, and $p$ is sampled between $0.1$ and $0.3$ with uniform probability.
This is done to introduce some variability in size and density in the dataset. Overall, $ER_{S}$ contains $600$ graphs.
We follow the same methodology for $ER_{L}$, except that $n$ is drawn uniformly between $300$ and $500$. Overall, $ER_{L}$ contains $200$ graphs.

\subsubsection{SuiteSparse matrix collection}

The SuiteSparse matrix collection\footnote{SuiteSparse matrix collection was formerly known as the University of Florida sparse matrix collection.} \cite{UFS} is a dataset of (sparse) matrices collected from a number of real-life applications, and is routinely used as benchmark for numerical linear algebra software.
Given a matrix $M$, we construct a non-oriented graph whose adjacency matrix corresponds exactly to the sparsity structure of $M$.
We only consider square matrices, and any non-symmetric matrix is transferred into symmetric by adding its transpose to it.

First, we select square matrices of size between $50$ and $500$.
This yields a dataset of $278$ graphs, which we denote by $SS_{S}$.
Similarly, we select square matrices of size between $1000$ and $2000$, and obtain a second dataset, denoted by $SS_{L}$, which contains $295$ graphs.

\subsection{Experimental settings}
\label{sec:experiments:settings}

Our experiments were conducted on a dual Intel Xeon Gold 6126@2.60GHz, 768BG RAM machine running Linux and equipped with Nvidia Tesla V100 GPUs.
Our code\footnote{The code repository as well as the instances are in the process of being made publicly available.} is written in Python 3.6, and we use Pytorch 0.4 for modeling and training GNNs.

\paragraph{Datasets.}
We split the $ER_S$ dataset into $\{training,\;validation,\; test\}$, each containing 200 graphs. Our GNN policy is trained and validated only with the $training$ and $validation$ set of $ER_S$, respectively. Then, we test the generalization performance of the trained model with the $test$ set of $ER_S$, $ER_L$, $SS_S$ and $SS_L$.

\paragraph{GNN setting.}
We apply the GNN architecture described in Section \ref{sec:learning:onpolicyil} with 2 layers. As initially the vertices of the graphs have no attribute, we initialize the feature of each vertex with the same value. Specifically, we take $h_v^0 = 1$, $\forall v \in V $. As a result, the encoding of each vertex only depends on the topological structure of its neighborhoods. The dimension of features in all layers is the same. For each layer, the weights are initialized from Xavier normal distribution \cite{glorot2010understanding} and we initialize the bias with zero. 

\paragraph{Performance metrics.}
For a finite dataset  $\mathcal{G}$, the first metric computes the average KL loss, given by 
\begin{align}
    \hat{\mathcal{L}}_{KL} = \frac{1}{\sum_{g \in \mathcal{G}} n_g} \sum_{g \in \mathcal{G}} \sum_{i =1}^{n_g} \mathcal{L}_{KL}\left(\pi_{\theta}(s_i), \pi_{MD}(s_i)\right),
\end{align}
where $n_g$ is the size of each graph $g \in \mathcal{G}$ and $\mathcal{L}_{KL}(\cdot)$ specifies the KL divergence between $\pi_{\theta}(s_i)$ and $\pi_{MD}(s_i)$ in state $s_i$ of $g$.

To measure the fill-in of a policy, the second metric computes the average number of fill-in per graph. For each graph $g \in \mathcal{G}$, we denote the total number of fill-in by $c_{fillin}^{g}$. Then, the average fill-in per graph is given by
\begin{align}
    \hat{C}_{fillin} = \frac{1}{|\mathcal{G}|} \sum_{g \in \mathcal{G}} c_{fillin}^{g}.
\end{align}

\begin{figure}[h!]
    \centering
    \includegraphics[width=1.0\textwidth]{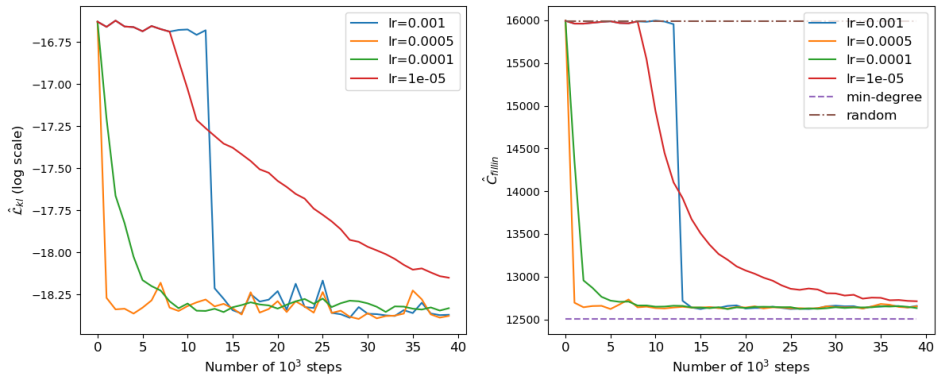}
    \caption{Validation results of imitation learning. We plot the average KL loss in $log$ scale (left) and the average fill-in per graph (right) on the validation set of $ER_S$. For fill-in, we compare GNN with minimum degree and random policy.}
    \label{fig:hypertuning}
\end{figure}

\paragraph{Training and validation.}
We train our GNN policy with Algorithm \ref{alg:the_alg}. At each epoch, we randomly shift the training set and sample single-graph mini batches.
For learning rate tuning, we experiment different learning rates from $10^{-5}$ to $10^{-3}$. The validation result is shown by plotting the average KL loss and the average fill-in per graph in Figure \ref{fig:hypertuning}. Observing that $10^{-4}$ yields fast and smooth convergence, we train the model with the learning rate of $10^{-4}$ for $20$ epochs. Moreover, we also observe a plateau effect for larger step size in Figure \ref{fig:hypertuning}, notably, sudden decrease with larger step size. This effect will be discussed in Section \ref{sec:further}.

\paragraph{Test.}
We test the generalization performance of trained GNN with four test sets as specified before. Addtionally, to evaluate the performance of GNN models at different stages of training, we first save the trained model 
at the end of each epoch. Then, we test the performance of each saved model on four test sets.

\subsection{Results}
\label{sec:experiments:results}

\begin{figure}[h!]
    \includegraphics[width=1.0\textwidth]{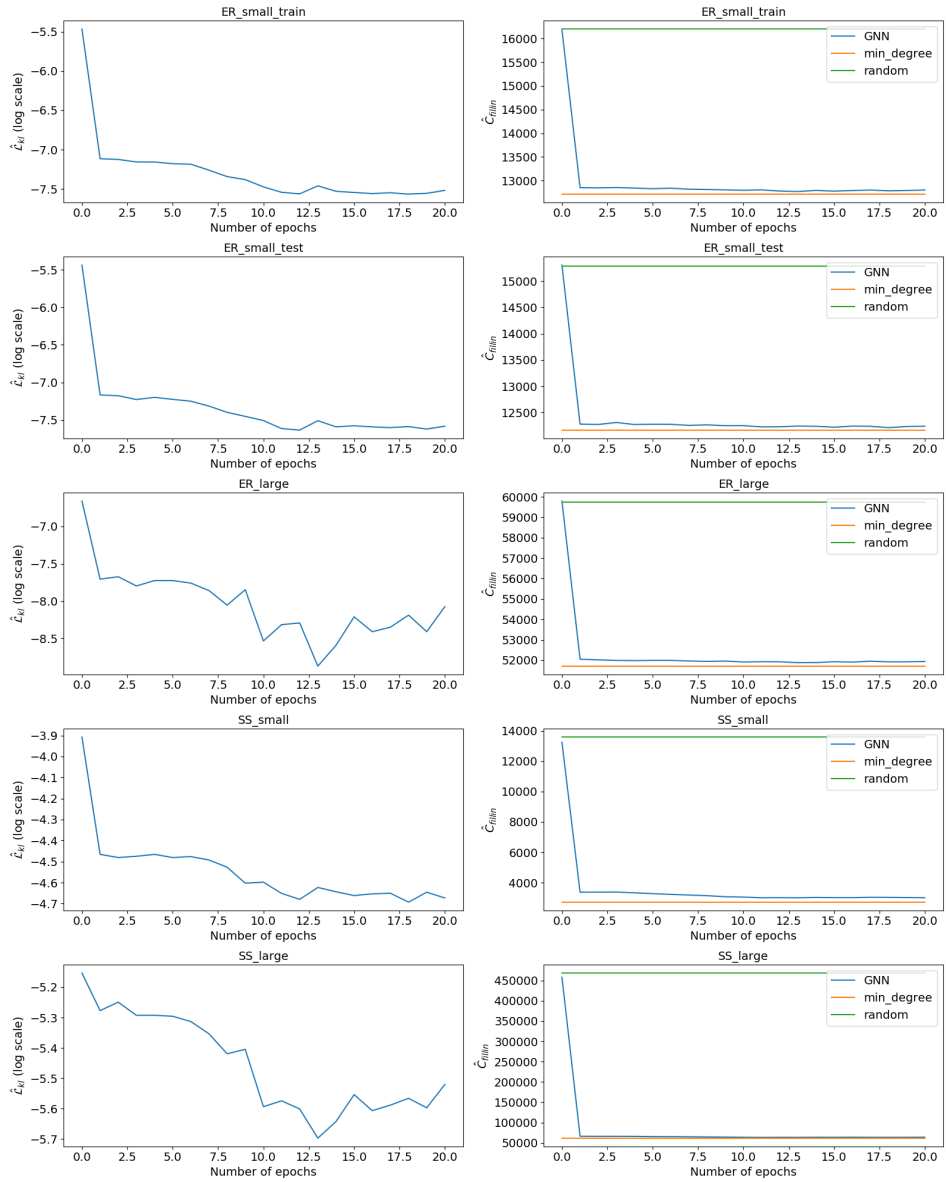}
    \caption{Test results of imitation learning. We plot the average KL loss in $log$ scale (left) and the average fill-in per graph (right) on the training set and four test sets. For fill-in, we compare our GNN with minimum degree and random policy. }
    \label{fig:res_train_per_epoch}
\end{figure}

In this section, we compare the predictive performance of GNN with the two metrics introduced in the previous section.
The results on the training set and four test sets are shown in Figure \ref{fig:res_train_per_epoch}.
Specifically, we plot the curves of two metrics over the entire training period (20 epochs), in order to compare the performance of GNN models at different stages of training.

From the results of the training set (shown in the first row of Figure \ref{fig:res_train_per_epoch}), we observe that the loss significantly decreases and stabilizes after about 15 epochs of training. Moreover, for fill-in, the GNN also matches the minimum degree heuristic. 

Comparing the results of different test sets, 
we observe that our GNN generalizes well, both to larger size graphs and to different distributions. First of all, loss curves of all datasets show same decreasing tendency over the entire training period, although the magnitude of values can be different across datasets. In terms of fill-in, we have similar and consistent results. Moreover, by comparing the curves in each row, we also observe a strong correlation between KL loss (i.e., how good we replicate the minimum degree expert), and the actual fill-in (which is only observed, we never learn anything from it). 

It is worth noting that, although the initial graphs can be i.i.d., the other states in the trajectory always depend on previous states and actions, which indicates the induced distribution of states depends on the behavior policy itself. 

Since we use the on-policy imitation learning approach, the loss is measured under the distribution of states induced by our GNN policy. As GNN model changes during training, this metric is actually measured under different distributions.  As a result, the decrease of loss over the training period (shown on the left side of Figure \ref{fig:res_train_per_epoch}) only shows a tendency that the GNN replicates the minimum degree expert better on an evolutionary distribution induced by itself. The predictive performance of the GNN still needs to be validated by the actual fill-in, which is precisely done on the right side of Figure \ref{fig:res_train_per_epoch}.

\section{Further discussion}
\label{sec:further}

We now seek to further explain the sharp drops in loss that were observed during training, e.g., in Figure \ref{fig:hypertuning}, and the stark correlation between imitation loss and fill-in.

To do so, we consider the following GNN with two layers:
\begin{align}
    x^{0}_{i} &= 1 & \forall i \in V,\\
    h^{1}_{i} &= \sum_{j \in \mathcal{N}(i)} w_{1} x^{0}_{j} & \forall i \in V,\\
    x^{1}_{i} &= Relu ( 1 + h^{1}_{i}) & \forall i \in V,\\
    h^{2}_{i} &= \sum_{j \in \mathcal{N}(i)} w_{2} x^{1}_{j} & \forall i \in V,\\
    x^{2}     &= Softmax(h^{2}),
\end{align}
where $w_{1}, w_{2} \in \mathbb{R}$ are the only two scalar parameters of the GNN, and $x^{0}, h^{1}, x^{1}, h^{2}, x^{2}$ are vectors of size $|V|$.
The input vector is $x^{0}$ with all coordinates equal to one and, by definition of $Softmax$, the coordinates of the output vector $x^{2}$ are all non-negative and sum to one.
Also note that, for every node $i$, since $x^{0}_{i}=1$, we have $h^{1}_{i} = w_{1} \delta(i)$.
It follows that, by setting $w_{1} = 0$, we get $x^{1}_{i} = Relu(1 + 0) = 1$, and then $h^{2}_{i} = w_{2} \times \delta(i)$.
Therefore, as $w_{2}$ approaches $-\infty$, $x^{2}$ becomes arbitrarily close to a minimum degree distribution.

\subsection{Landscape of the loss function}
\label{sec:further:loss}

\begin{figure}[h]
    \centering
    \includegraphics[width=0.8\textwidth]{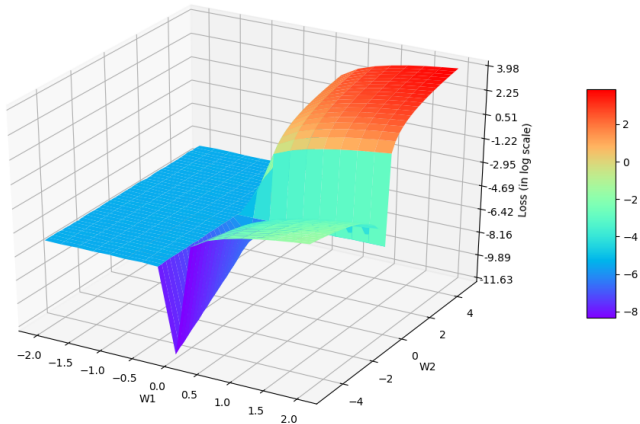}
    \caption{Landscape of the expected average KL loss (in log scale). For each $(w_{1}, w_{2})$, we plot the expected average KL loss, estimated over the training set.}
    \label{fig:landscape_loss}
\end{figure}

We begin by plotting the landscape of the expected average KL loss, evaluated on the training set.
This landscape is represented in Figure \ref{fig:landscape_loss}.
Although this corresponds to a simpler model than the one that yielded the results in Section \ref{sec:experiments:results}, it gives several insights into the behavior during training.

First, as expected, the average loss is minimized when $w_{1} = 0$ and $w_{2}$ goes to $-\infty$.

Second, we observe that in the $w_{1} \leq -1$ region, the average loss is flat.
This region actually corresponds to the $Relu$ of the first layer being inactive.
Indeed, we have $h^{1}_{i} = w_{1} \delta(i)$, therefore, when $w_{1} \leq -1$, we automatically get $1 + h^{1}_{i} \leq 0$, which yields $x^{1}_{i}=0$.
Consequently, the output of the GNN is a uniform distribution on the nodes of the graph, i.e., we obtain $x^{2}_{i} = \frac{1}{n}$ for each node $i \in \{ 1, ..., n\}$.

Third, observe that the landscape of the loss function displays fairly flat regions, which tend to be separated by sharp drops in the objective, e.g., around the $w_{1} =0$ region.
This landscape most likely explains the shapes of the training curves in Figure \ref{fig:hypertuning}, which displayed flat progression followed by sharp drops in the loss.
Whether such behavior would carry out in larger dimensions remains an open question.

\subsection{Landscape of the fill-in}
\label{sec:further:fill}

\begin{figure}[h]
    \centering
    \includegraphics[width=0.8\textwidth]{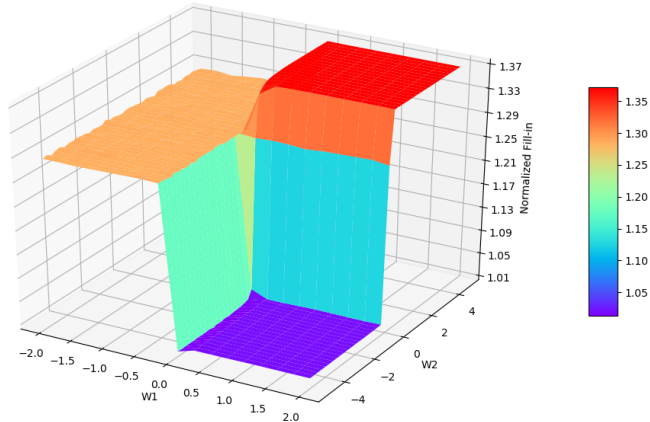}
    \caption{Landscape of the normalized total fill-in.
    For each $(w_{1}, w_{2})$, we plot the average total fill-in of the corresponding GNN policy, divided by the expected total fill-in of the minimum degree heuristic. Both expectations are estimated over the training set.}
    \label{fig:landscape_fill}
\end{figure}

We then plot the landscape of the expected total fill-in in Figure \ref{fig:landscape_fill}, also evaluated on the training set.
While this gives us an insight into how fill-in correlates with the KL loss, let us formally restate that fill-in is never used during the training process.
In particular, no gradient information is ever inferred from fill-in.

Fist, unsurprisingly, similar to Figure \ref{fig:landscape_loss}, here we observe a flat landscape in the $w_{1} \leq -1$ region.
Recall indeed that setting $w_{1} \leq -1$ means the GNN's output reduces to a uniform policy.
Second, the region $w_{1} \geq 0, w_{2} \geq 0$ displays high fill-in.
This is not surprising either since this region essentially yields policies that select nodes with high degree, which is naturally detrimental to fill-in.

A third and more remarkable observation is the flat valley in the region $w_{1} \geq 0, w_{2} \leq 0$.
While we know that the GNN policy converges to minimum degree when $w_{1} = 0$ and $w_{2}$ takes large negative values, the plots in Figure \ref{fig:landscape_fill} show that, when it comes to fill-in, the magnitude of $w_{2}$ does not matter as much.

Fourth and last, the minimum degree policy appears to be a minimizer of the expected total fill-in, among the set of policies that are representable by the class of GNN at hand.
Although we cannot extrapolate to larger classes of models, nor to other datasets of graphs, this last observation has consequences if one were to train a GNN to minimize fill-in.
Specifically, one would \emph{need} models with higher representation power to achieve better fill-in than the minimum degree algorithm.

\section{Conclusion}
\label{sec:conclusion}

In this work, we have considered chordal extensions and graph elimination as major factors for devising sparsity-exploiting techniques for optimization algorithms.
We have argued that, although effective heuristics to perform graph elimination (an NP-complete task) exist, there is no definitive understanding of the effect of the obtained chordal extension on the optimization algorithm using the final graph representation.

For this reason, we have followed the current research trend of looking at Combinatorial Optimization tasks by using a Machine Learning lens and we have devised a framework for learning elimination rules yielding high-quality chordal extensions.
As a first building block of the learning framework, we have proposed an on-policy imitation learning scheme that mimics the elimination ordering provided by the (classical) minimum degree rule. 

The results have shown that our on-policy imitation learning approach is effective in learning the minimum degree policy and, consequently, produces graphs with desirable fill-in characteristics.
In addition, the learned policy displays remarkable generalization performance, a desirable behavior since it allows to speed-up the learning process by training on smaller problems.

Finally, we identify two main research avenues for subsequent developments.
On one hand, while GNNs are a good model prior for combinatorial problems over graphs, enlarging their representation power, for instance to represent hypernodes or to model multiple eliminations, will likely be key to handling practical tasks.
On the other hand, the next logical step will be to learn elimination rules that explicitly address the performance of practical optimization algorithms, in conjunction with reinforcement learning-based approaches.
In that regard, our future work will investigate chordal decomposition specially tailored to SDP optimization problems.

\bibliographystyle{unsrt}  
\bibliography{refs}   






\end{document}